\pdfoutput=1

\documentclass[11pt]{article}

\usepackage[preprint]{acl}

\usepackage{times}
\usepackage{latexsym}

\usepackage[T1]{fontenc}

\usepackage[utf8]{inputenc}

\usepackage{microtype}

\usepackage{inconsolata}
\usepackage[most]{tcolorbox}
\usepackage{enumitem}
\usepackage{xspace}
\usepackage{geometry}
\usepackage{setspace}
\usepackage[]{multirow}
\usepackage{multicol}

\usepackage{makecell}
\usepackage[]{booktabs}
\usepackage{subcaption}
\usepackage{array}
\usepackage{pifont}
\usepackage{booktabs}
\usepackage{pifont}
\usepackage[]{adjustbox}
\usepackage[]{amssymb}

\usepackage{graphicx}
\usepackage{algorithm}
\usepackage{algpseudocode}
\usepackage{cleveref}
\usepackage{tcolorbox}
\usepackage[utf8]{inputenc}

\newcommand{\cmark}{{\ding{51}}} 
\newcommand{\xmark}{\textcolor{red}{\ding{55}}}  
\newcommand{\qmark}{\textcolor{orange}{\textbf{?}}}  
\newcommand{\sys}{\textsc{VeriMAP}\xspace}

\newcommand{\saexecutor}{ReAct\xspace}
\newcommand{\saplanner}{ReAct\xspace}
\newcommand{\noveri}{MAP\xspace}
\newcommand{\generic}{MAP-V\xspace}
\newcommand{\genericOneR}{MAP-V-1it\xspace}
\newcommand{\sysOneR}{\sys-1it\xspace}

\tcbset{
    userstyle/.style={
        enhanced,
        colback=white,
        colframe=black,
        colbacktitle=gray!20,
        coltitle=black,
        rounded corners,
        sharp corners=north,
        boxrule=0.5pt,
        drop shadow=black!50!white,
        attach boxed title to top left={
            xshift=-2mm,
            yshift=-2mm
        },
        boxed title style={
            rounded corners,
            size=small,
            colback=gray!20
        },
        fontupper=\footnotesize,
        left=1mm,
        right=1mm,
        top=2mm,
        bottom=1mm
    },
    jailbreakstyle/.style={
        enhanced,
        colback=white,
        colframe=red,
        colbacktitle=red!40,
        coltitle=black,
        rounded corners,
        sharp corners=north,
        boxrule=0.5pt,
        drop shadow=red!50!white,
        attach boxed title to top left={
            xshift=-2mm,
            yshift=-2mm
        },
        boxed title style={
            rounded corners,
            size=small,
            colback=red!20
        },
        fontupper=\footnotesize,
        left=1mm,
        right=1mm,
        top=2mm,
        bottom=1mm
    },
    jailbreakstyleres/.style={
        enhanced,
        colback=white,
        colframe=red,
        colbacktitle=red!40,
        coltitle=black,
        rounded corners,
        sharp corners=north,
        boxrule=0.5pt,
        drop shadow=red!50!white,
        attach boxed title to top right={
            xshift=-2mm,
            yshift=-2mm
        },
        boxed title style={
            rounded corners, 
            size=small,
            colback=red!0
        },
        fontupper=\footnotesize,
        left=1mm,
        right=1mm,
        top=2mm,
        bottom=1mm
    },
    myreplyborderstyle/.style={
        enhanced,
        colback=white,
        colframe=black,
        colbacktitle=red!40,
        coltitle=black,
        rounded corners,
        sharp corners=north,
        boxrule=0.5pt,
        drop shadow=black!50!white,
        attach boxed title to top right={
            xshift=-2mm,
            yshift=-2mm
        },
        boxed title style={
            rounded corners, 
            size=small,
            colback=red!0
        },
        fontupper=\footnotesize,
        left=1mm,
        right=1mm,
        top=2mm,
        bottom=1mm
    },
    replystyleg/.style={
        enhanced,
        colback=blue!0,
        colbacktitle=black,
        colframe=black,
        coltitle=black,
        boxrule=1pt,
        drop shadow=black!50!,
        rounded corners,
        sharp corners=north,
        attach boxed title to top right={
            xshift=-2mm,
            yshift=-2mm
        },
        boxed title style={
            rounded corners,
            size=small, 
            colback=blue!0,
        },
        fontupper=\footnotesize,
        left=1mm,
        right=1mm,
        top=2mm,
        bottom=1mm
    },
    replystyler/.style={
        enhanced,
        colback=red!15,
        colframe=black,
        colbacktitle=red!40,
        coltitle=black,
        boxrule=0.5pt,
        drop shadow=black!50!white,
        rounded corners,
        sharp corners=north,
        attach boxed title to top right={
            xshift=-2mm,
            yshift=-2mm
        },
        boxed title style={
            rounded corners,
            size=small,
        },
        fontupper=\footnotesize,
        left=1mm,
        right=1mm,
        top=2mm,
        bottom=1mm
    },
    replystylew/.style={
        enhanced,
        colback=purple!5,
        colframe=black,
        colbacktitle=pink!40,
        coltitle=black,
        boxrule=0.5pt,
        drop shadow=black!50!white,
        rounded corners,
        sharp corners=north,
        attach boxed title to top right={
            xshift=-2mm,
            yshift=-2mm
        },
        boxed title style={
            rounded corners,
            size=small,
            colback=pink!60
        },
        fontupper=\footnotesize,
        left=1mm,
        right=1mm,
        top=2mm,
        bottom=1mm
    }
}

\newtcolorbox{userquery}[1][]{
    userstyle,
    title=Prompt,
    #1
}

\newtcolorbox{llmreply-g}[1][]{
    replystyleg,
    title=Response,
    #1
}

\newtcolorbox{llmreply-r}[1][]{
    replystyler,
    title=Response,
    #1
}

\newtcolorbox{mybox}[2][]{
    replystyler,
    title=#2,
    #1
}
\newtcolorbox{myboxw}[2][]{
    replystylew,
    title=#2,
    #1
}

\newtcolorbox{myboxg}[2][]{
    replystyleg,
    title=#2,
    #1
}

\newtcolorbox{myuser}[2][]{
    userstyle,
    title=#2,
    #1
}

\newtcolorbox{myjailbreak}[2][]{
    jailbreakstyle,
    title=#2,
    #1
}

\newtcolorbox{myreplyborder}[2][]{
    myreplyborderstyle,
    title=#2,
    #1
}

\tcbset{
  myprompt/.style={
    enhanced,
    breakable,
    colback=blue!5!white, 
    colframe=blue!70!black,
    coltitle=white,
    fonttitle=\bfseries,
    title style={fill=blue!70!black},
    boxrule=0.8pt,
    arc=2mm,
    top=4pt, bottom=4pt, left=6pt, right=6pt,
    drop shadow
  }
}

\title{Verification-Aware Planning for Multi-Agent Systems}

\author{
  Tianyang Xu\thanks{Work done during internship at Megagon Labs.}\\
  Purdue University, USA\\
  \texttt{xu1868@purdue.edu}
  \And
  Dan Zhang\\
  Megagon Labs, USA\\
  \texttt{dan\_z@megagon.ai}
  \AND
  Kushan Mitra\\
  Megagon Labs, USA\\
  \texttt{kushan@megagon.ai}
  \And
  Estevam Hruschka\\
  Megagon Labs, USA\\
  \texttt{estevam@megagon.ai}
}

\begin{document}
\maketitle
\begin{abstract}
Large language model (LLM) agents are increasingly deployed to tackle complex tasks, often necessitating collaboration among multiple specialized agents. However, multi-agent collaboration introduces new challenges in planning, coordination, and verification. Execution failures frequently arise not from flawed reasoning alone, but from subtle misalignments in task interpretation, output format, or inter-agent handoffs. To address these challenges, we present \sys{}, a framework for multi-agent collaboration with verification-aware planning. The \sys{} planner decomposes tasks, models subtask dependencies, and encodes planner-defined passing criteria as subtask verification functions (VFs) in Python and natural language. We evaluate \sys{} on diverse datasets, demonstrating that it outperforms both single- and multi-agent baselines while enhancing system robustness and interpretability. Our analysis highlights how verification-aware planning enables reliable coordination and iterative refinement in multi-agent systems, without relying on external labels or annotations.
\end{abstract}

\section{Introduction}\label{sec:intro}

Large language model (LLM) agents are increasingly deployed to solve diverse and complex tasks. Beyond single-agent settings, multiple agents can be orchestrated to collaborate on a shared objective~\citep{zhu-etal-2025-multiagentbench, cemri2025multiagentllmsystemsfail, li2023camelcommunicativeagentsmind}, enabling workflows that combine the complementary strengths of different models and tools. 

For example, a reasoning-oriented agent can interpret and plan around the overall goal, determine suitable task granularity and instructions, and coordinate subtasks across agents—delegating bounded actions to lightweight executors and leveraging specialized tools for retrieval or computation.
This modularity creates opportunities for improved efficiency, interpretability, and control in AI systems, as practitioners can configure and combine models with tailored capabilities for planning, execution, and verification.

A central component of multi-agent collaboration is planning. Effective planning requires understanding the overall task, decomposing it into subtasks, modeling dependencies, and assigning responsibilities to agents with different capabilities. Often, plans are represented as directed acyclic graphs (DAGs)~\citep{prasad-etal-2024-adapt,sung2025verilahumancenteredevaluationframework,erdogan2025planandactimprovingplanningagents}, where the output of one agent becomes the input to another. This structure enables scalable problem-solving but introduces fragility: the workflow depends on smooth communication and accurate handoffs.

In multi-agent systems, agents fail in diverse ways. Verification is commonly used to monitor performance, typically checking factual accuracy, logical consistency, or final answer quality~\citep{miao2024selfcheck,hao-etal-2025-large,grigorev2024verifyllm}. However, failures can range from obvious to subtle: an agent may (i) produce no result, (ii) output in the wrong format, breaking downstream consumption, or (iii) produce results that appear reasonable but diverge from what the planner or downstream agents expect. For example, in a document summarization pipeline, a ``section extractor'' might return raw text instead of structured JSON, or misinterpret ``summarize key claim'' as ``summarize all paragraphs'', causing downstream failures. These latter cases show that \textbf{execution failure is context-dependent} — correctness alone is insufficient if outputs violate plan expectations. Existing verification rarely captures such interpretation or handoff failures~\citep{stechly2024selfverificationlimitationslargelanguage,sung2025verilahumancenteredevaluationframework}.

To address these limitations, we propose \sys{}, a framework that integrates planning and verification in a contextualized manner. The planner decomposes tasks, models dependencies, and specifies requirements for each agent’s output. These requirements are encoded as verification functions (VFs), which distill global context into local checks, allowing verifiers to operate independently while ensuring outputs satisfy both local and global expectations. Analogous to good project management, tasks are assigned with clear acceptance criteria, and handoffs follow well-defined ``passing standards''. With these explicit requirements, agents gain a precise understanding of what is expected, can self-refine when their outputs do not conform, and can offload global reasoning to the planner while focusing their efforts on localized subtasks. This allows smaller models to participate effectively, avoids redundant correction of upstream errors, and ultimately improves system robustness.

\begin{figure*}[ht]
    \centering
    \includegraphics[width=0.95\linewidth]{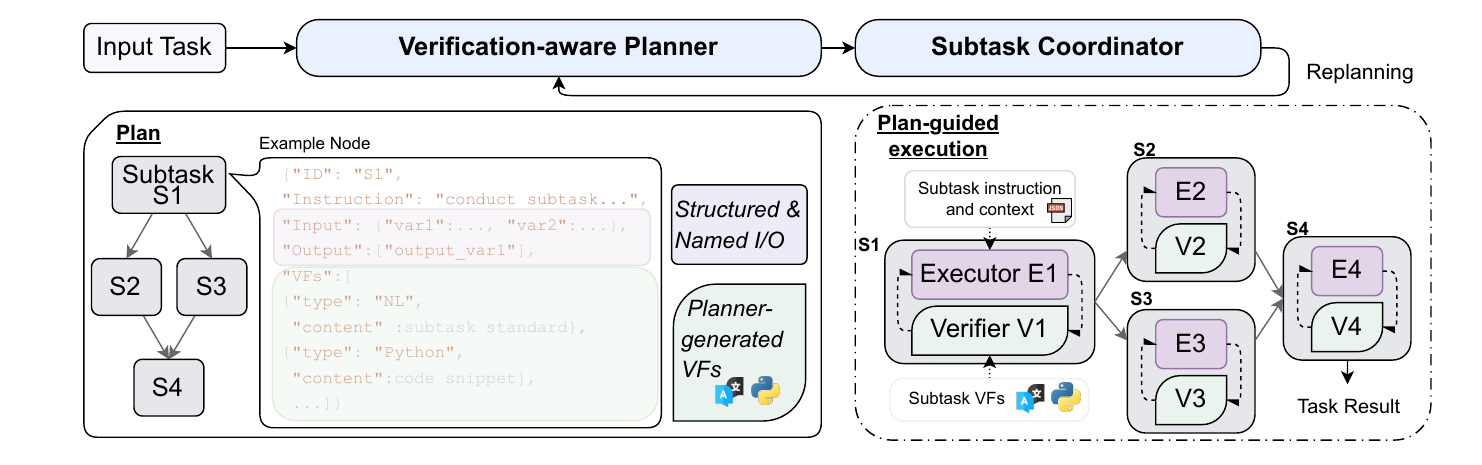}
    \caption{Overview of \sys. \sys integrates planning and verification by decomposing tasks into a DAG of subtasks, where the planner specifies Structured and Named  I/O and Verification Functions (VFs) at subtask-level. Executors produce JSON outputs verified by paired VFs, while the coordinator manages contexts, retries, and replanning to ensure reliable final results.}
    \label{fig:diagram}
\end{figure*}

In this paper, we make the following contributions:
\begin{itemize}
    \item We introduce \sys{}, a system for multi-agent collaboration with verification-aware planning. The planner decomposes tasks, models dependencies, and specifies requirements for each agent’s output using Structured I/O and per-subtask VFs in Python and natural language, which distill global context into local checks. This enables agents to self-refine, share context efficiently, focus on localized subtasks, and improve overall robustness.
    \item We evaluate \sys{} on five datasets covering mathematics, coding, and question answering. It outperforms single- and multi-agent baselines, particularly on harder datasets (up to 4.05\% on BigCodeBench-Hard and 9.8\% on Olympiads, compared with the next-best tool-enabled ReAct agents), and our analysis of cost, VF characteristics, and a case study demonstrates how verification-aware planning improves robustness, and enables iterative refinement.
\end{itemize}

\section{\sys}\label{sec:methods}

We propose \sys, a multi-agent LLM system integrating planning and verification. As illustrated in \Cref{fig:diagram}, the system consists of four key modules: \textbf{Planner}, \textbf{Executor}, \textbf{Verifier}, and \textbf{Coordinator}.

\subsection{Verification-Aware Planner}

The Verification-Aware Planner lies at the core of \sys, responsible for decomposing complex tasks into executable subtasks while simultaneously generating \textbf{Verification Functions} (VFs) for each subtask. Like conventional multi-agent planners, it constructs a directed acyclic graph (DAG) in which nodes represent subtasks and edges capture model dependencies. Each node is assigned a unique identifier and contains the subtask instruction, its dependency edges, and metadata that regulates both context sharing and dependency resolution.

\paragraph{Planning for Information Flow} In addition to planning subtask decomposition and dependencies, the \sys planner also orchestrates message passing across agents. To ensure a precise and transparent flow of information, \sys enforces: (i) Structured I/O, requiring that agent inputs and outputs adhere to well-defined formats, such as JSON objects; and (ii) Named Variables, requiring that each I/O object has a unique, consistent name across all nodes in the plan. Each node thus receives an input object consisting of \texttt{(name: value)} pairs, and the plan explicitly defines the expected outputs along with their variable names. This guarantees that downstream agents can reliably reference and consume upstream outputs, reducing execution ambiguity.

\vspace{-2mm}
\paragraph{Planner-Defined Subtask Verification} A second key contribution of the planner is the generation of \textbf{subtask-specific verification functions}. The planner’s structured and consistently named I/O provides a rigorous foundation for verification, supporting the generation of functions aligned with each subtask’s expected structure and semantics. For each node, the planner specifies precise criteria that determine whether an execution result is correct by generating verification functions (VFs) for that subtask. This design allows a verifier to focus on evaluating these planner-generated VFs, rather than reasoning about the overall task or other subtasks, effectively offloading much of the reasoning burden from the verifier to the planner.

For each subtask, the \sys planner generates VFs of two categories based on the task requirements.
\begin{itemize}
    \item \textbf{Python VFs.} For tasks with well-defined functional or structural requirements, the planner generates self-contained Python assertions. These validate output type, format, and correctness against specified criteria. Being programmatic, they provide deterministic and reproducible guarantees.

    \item \textbf{Natural language VFs.} For tasks requiring semantic or open-ended judgments (e.g., summaries, natural language answers), the planner generates instructions in natural language to guide a verifier agent. These complement the strictness of Python tests with broader semantic coverage.
\end{itemize}

\subsection{Executor}

The \textbf{Executor} is responsible for solving its assigned subtask using planner-generated instructions and necessary context (output of upstream nodes) and then produces structured outputs that can be passed to Verifiers and downstream Executors. To limit scope and reduce reasoning complexity, Executors are not exposed to the original global task. We implement the Executor using LLMs with general-purpose agent paradigms, such as \textit{ReAct}-style agents~\citep{yao2023react} and tool-calling agents, where planner-generated instructions are provided as task prompts. The Executor is capable of making lightweight decisions and invoking external tools as needed. While this work assumes general-purpose Executors, \sys can be easily extended to incorporate specialized agents (e.g., an information extraction agent or an NL2SQL agent) if available. Since each Executor typically handles a simple, well-scoped subtask, it can rely on smaller and more cost-efficient models; in our experiments, we use \texttt{gpt-4o-mini}. Additional implementation details are provided in Appendix~\ref{sec:executor_implementation}.

\subsection{Verifier}\label{sec:verifier}

As discussed earlier, the plan specifies a set of predefined verification functions (VFs) that capture the expected behavior of each subtask executor. After execution, the Verifier evaluates all VFs against the structured output of the Executor. For natural language VFs, it issues LLM calls with \texttt{(executor\_output, VF)} pairs; for Python-based VFs, it invokes a Python interpreter. The Verifier then aggregates the results from all VFs to make a final decision. In this paper, we adopt a strict logical \textsc{AND} strategy, where a subtask is marked as failed if any VF fails. In practice, softer aggregation methods could weight VFs differently according to their importance; we leave such extensions to future work. When a VF fails, the Verifier collects signals to guide retries: LLM-based VFs return explanations for the failure, while Python-based VFs provide error traces. Further implementation details are given in Appendix~\ref{sec:verifier_implementation}.

\subsection{Coordinator}

The \textbf{Subtask Coordinator} serves as the central orchestrator of multi-agent task execution, following the task plan (represented as a DAG) to support reliable and adaptive execution. Its responsibilities focus on task sequencing, context management, execution and verification monitoring, error handling, and dynamic replanning.

\paragraph{Task Sequencing and Context Management}

The coordinator first determines a topological order of task nodes to respect dependencies. Before executing each node, it compiles a context that includes outputs from prior tasks. This ensures that subtasks receive all necessary inputs and maintain consistency across dependent operations.

\paragraph{Execution and Verification Management}

For each node, the coordinator executes the task within a controlled retry loop, invoking the executor with proper context and invoking the verifier to check the executor output. If the verification passes, the coordinator proceeds to the next node in the plan sequence. If the verification fails, the coordinator retries the execution-verification loop up to a maximum number of attempts (defaulted to 3), updating the context with previous execution results to guide subsequent retries.

\paragraph{Error Handling and Replanning}
If a node fails verification after all retries, the coordinator will collect execution traces and trigger replanning to generate a revised task plan. Execution restarts from the top of the new plan. This approach adapts to failures, prevents the propagation of errors, and improves overall task correctness. Iteration limits (defaulted to 5) ensure termination, either with a successful final result or a failure report if maximum replanning cycles are exceeded. Replanning follows the same instructions from the \sys planner, adding only minimal guidance to inspect the execution trace and generate a new plan (details in \Cref{sec:planner_prompts}).

The complete coordinator pseudocode (\Cref{alg:task-coordinator-multi-vf}) and additional implementation details are provided in Appendix~\ref{sec:coordinator_implementation}.
\begin{table*}[ht]
\centering
\small
\begin{tabular}{lccccc}
\toprule
\textbf{Method} & \multicolumn{1}{c}{\textbf{RAG}} & \multicolumn{2}{c}{\textbf{Programming}} & \multicolumn{2}{c}{\textbf{Math}} \\
\cmidrule(lr){2-2}\cmidrule(lr){3-4}\cmidrule(lr){5-6}
 & MultiHopRAG & HumanEval & BigCodeBench-Hard & GSM8K & Olympiads \\
\midrule
\saexecutor (\texttt{gpt-4o-mini}) & 61.20\% & 81.10\% & 27.03\% & 90.00\% & 25.00\% \\
\saplanner  (\texttt{gpt-4.1})& 68.40\% & 90.24\% & 36.49\% & 91.80\% & 40.80\% \\
\noveri      & 67.00\% & 78.88\% & 28.38\% & 57.20\% & 21.40\% \\
\generic     & 77.60\% & 88.96\% & 28.38\% & 87.00\% & 29.00\% \\
\midrule
\sys  (ours) & \textbf{78.20\%} & \textbf{93.92\%} & \textbf{40.54\%} & \textbf{93.60\%} & \textbf{41.20}\% \\
\midrule

\genericOneR  & 73.20\% & 86.50\% & 28.38\% & 82.00\% & 23.60\% \\
\sysOneR     & 75.40\% & 91.22\% & 31.08\% & 89.20\% & 32.00\% \\
\bottomrule
\end{tabular}
\caption{Accuracy of single- and multi-agent baselines, \sys, and their variants. \texttt{-1it} denotes ablation variants limited to a single iteration without replanning. All multi-agent systems use \texttt{gpt-4.1} as the planner and \texttt{gpt-4o-mini} as the executor. The best results are shown in \textbf{bold}.}
\label{tab:main_results}

\end{table*}

\section{Evaluation}

\subsection{Setup}
\paragraph{Evaluation Datasets}
We evaluate \sys on five datasets (full details in Appendix~\ref{sec:dataset_details} and Table~\ref{tab:datasets}) spanning Question Answering (QA), Programming, and Math tasks:

\begin{itemize}[noitemsep, topsep=0.5pt]
    \item \textbf{MultiHopRAG}~\citep{tang2024multihoprag}, a question-answering dataset designed to require multi-hop retrieving and reasoning to find the correct answer;
    \item  \textbf{HumanEval}~\citep{chen2021evaluating}, a human-written programming dataset to evaluate the functional correctness of generated code;
    \item \textbf{BigCodeBench-Hard}~\citep{zhuo2024bigcodebench}, a challenging subset of BigCodeBench with complex, real-world programming tasks;
    \item \textbf{GSM8K}~\citep{cobbe2021training}, a high-quality grade school math problems dataset;
    \item \textbf{Olympiads}~\citep{li2024numinamath}, a curated collection of challenging problems from various mathematics olympiads.

\end{itemize}

\paragraph{Tools}
We equip \sys and baselines with tools tailored to the task type:
for Programming, we provide a sandboxed environment with tools for file operations (create, edit, delete) and for executing bash commands, Python scripts, and pytest tests.
For Math tasks, the agent has a sandboxed Python executor, a simple arithmetic calculator, and a computer algebra system-enabled calculator.
For QA tasks, we provide a specialized search tool built on Sentence Transformers, which supports pagination and conditional filtering of articles by attributes like title, author, category, date, source, and content.

\paragraph{Baselines} 
We first compared \sys against \emph{ReAct} agents instantiated with different backbone LLMs. Each agent was given access to the full set of tools available in our system. The backbone models were chosen to mirror the stronger and weaker models used in \sys's planner and other components (i.e., executor, verifier), respectively.

For multi-agent baselines, we implemented two variants, denoted as \textbf{MAP} and \textbf{MAP-V}.  
\begin{itemize}[leftmargin=*]
    \item \textbf{MAP} The Multi-Agent-Planning (MAP) baseline follows AOP\footnote{Unlike AOP, functionalities such as math, search, and code are abstracted as tools that executors can flexibly invoke, rather than being modeled as independent agents directly addressed in the plan.}~\cite{li2025agentoriented} setup, where multiple agents collaborate through planning  without verification. Each agent is equipped with tool calling and ReAct-style reasoning capabilities.
    
    \item \textbf{MAP-V} adds generic LLM verifiers to the MAP baseline, following the setup of VeriLA~\cite{sung2025verilahumancenteredevaluationframework} and SelfCheck~\cite{miao2024selfcheck}\footnote{Note that SelfCheck performs step-wise verification of chain-of-thought reasoning, which is comparable but not identical to verifying subtasks through a separate agent.}. The verifier is implemented as a \textbf{ReAct agent} with access to all available tools. This allows it to make independent decisions about which tools to invoke when verifying subtask outputs, given the corresponding instruction and context. This baseline can be viewed as a simplified variant of \sys, generating a single NL-based verification function that treats the subtask instruction itself as the verification criterion.
\end{itemize}

More details can be found in Appendix \ref{sec:appendix_baselines}.

\paragraph{Evaluation Metrics}
Across all datasets, we report the outcome accuracy as our primary evaluation metric. For the math and QA tasks, we determine the correctness of a generation using a heuristic-based criterion. For the programming tasks, a solution is considered correct if it passes all the accompanying test cases on the first attempt (\texttt{pass@1}). All results are averages of three runs.

\paragraph{Implementation Details}
For the underlying LLMs, we utilize \texttt{gpt-4.1} for the planner and \texttt{gpt-4o-mini} for the executor, unless otherwise specified. The temperature is set to $1.0$ and top\_p is $1.0$. Additional implementation details are provided in Appendix~\ref{sec:implementation}.

\subsection{Results}

Table~\ref{tab:main_results} reports accuracy across five benchmarks, showing that \sys consistently achieves the best performance on all tasks.
\paragraph{Comparison with single-agent methods}
\sys outperforms the strong tool-enabled \saplanner agent (backed by the same reasoning model \texttt{gpt-4.1}) by a significant margin of +9.8\% on MultiHopRAG and +4.05\% on the challenging BigCodeBench-Hard benchmark, demonstrating the effectiveness of planning and verification in complex reasoning tasks.

\paragraph{Comparing with multi-agent systems}
When multi-agent coordination is ineffective, single-agent solutions powered by strong models can outperform them (ReAct with \texttt{gpt-4.1} outperforms \noveri on all five datasets and beats \generic on four). In contrast, \sys effectively leverages planner-generated VFs to enhance performance, achieving notable gains of +12.2\% on Olympiads and +12.16\% on BigCodeBench-Hard.

\paragraph{Effect of replanning} 
To investigate the impact of replanning and the role of VFs in enhancing replanning quality, we report single-iteration variants (\sysOneR and \genericOneR) that do not perform replanning at the bottom of \Cref{tab:main_results}. As expected, replanning has a clearly positive effect. However, even without replanning, \sysOneR outperforms the \generic baseline with replanning on most benchmarks as well as the \saexecutor baseline with \texttt{gpt-4o-mini}. More importantly, when replanning is applied, \sys achieves substantial gains of +9.46\% on BigCodeBench-Hard and +9.2\% on Olympiads benchmarks, whereas replanning provides limited benefit for \generic (0\% and +5.39\%, respectively). These results indicate that planner-generated VFs capture more informative runtime signals, which significantly enhance performance during the replanning phase. Additional statistics on the average number of iterations are provided in \Cref{app:iteration_counts}.

\paragraph{Results on other LLMs} 
\begin{table*}[t]
\centering
\small
\setlength{\tabcolsep}{6pt}
\begin{tabular}{lccccc}
\toprule
\textbf{Method} & \multicolumn{1}{c}{\textbf{RAG}} & \multicolumn{2}{c}{\textbf{Programming}} & \multicolumn{2}{c}{\textbf{Math}} \\
\cmidrule(lr){2-2}\cmidrule(lr){3-4}\cmidrule(lr){5-6}
 & MultiHopRAG & HumanEval & BigCodeBench-Hard & GSM8K & Olympiads \\
\midrule
\multicolumn{1}{l}{\textit{Single-agent baselines}} & & & & & \\
\saplanner  (\texttt{o3})& 31.80\% & 95.45\% & 39.19\% & \textbf{95.60\%} & 58.60\% \\
\saplanner (\texttt{claude-sonnet-4})  & \textbf{84.40\%} & \textbf{95.68\%} & 27.84\% & 61.40\% & 34.20\% \\
\midrule
\multicolumn{1}{l}{\textit{\texttt{o3} as planner}} & & & & & \\
\noveri      & 52.00\% & 88.41\% & 24.32\% & 57.80\% & 47.80\% \\
\generic     & 65.80\% & 87.80\% & 28.38\% & 89.60\% & 56.90\% \\
\sys         & 55.80\% & 89.04\% & \textbf{40.54\%} & \textbf{95.60\%} & \textbf{68.40\%} \\
\midrule
\multicolumn{1}{l}{\textit{\texttt{claude-sonnet-4} as planner}} & & & & & \\
\noveri      & 74.20\% & 84.00\% & 20.27\% & 50.40\% & 26.05\% \\
\generic     & 79.00\% & 86.40\% & 22.22\% & 84.60\% & 38.07\% \\
\sys        & 64.60\% & 87.60\% & 32.39\% & \textbf{95.60\%} & 47.69\% \\
\bottomrule\end{tabular}
\caption{Accuracy of single- and multi-agent baselines and \sys on other LLMs. All multi-agent systems use \texttt{gpt-4o-mini} as the executor unless otherwise stated. The best results are shown in \textbf{bold}.}
\label{tab:main_results_o3_claude}

\end{table*}
Additionally, we experiment with \texttt{o3} and \texttt{claude-sonnet-4} as planners to assess the generality of \sys (\Cref{tab:main_results_o3_claude}). Overall, the results further validate the effectiveness of the verification-aware planning design: \sys with a reasoning-oriented planner achieves the strongest performance on three datasets. Notably, on the challenging Olympiads benchmark, \sys attains substantial improvements of +9.8\% over ReAct with \texttt{o3} and +34.2\% over ReAct with \texttt{claude-sonnet-4}. At the same time, the analysis reveals that different models excel in different domains. For instance, ReAct with \texttt{o3} performs poorly on MultiHopRAG (31.8\%), whereas \texttt{claude-sonnet-4} achieves markedly higher accuracy (84.4\%), yet \texttt{o3} outperforms \texttt{claude-sonnet-4} on math-related tasks. This discrepancy likely stems from model specialization: \texttt{claude-sonnet-4} is optimized for tool-use behaviors such as retrieval, core to question answering in MultiHopRAG, while \texttt{o3} demonstrates stronger formal reasoning capabilities essential for mathematical problem-solving. Furthermore, when a model already performs exceptionally well on datasets like MultiHopRAG or HumanEval (potentially due to training exposure), generating high-quality plans may become more difficult than directly solving the problem, as evidenced by the observed performance drops of multi-agent solutions on these tasks.

\paragraph{Cost Analysis}
\begin{figure*}[t]
    \centering
    \small
    
        \begin{subfigure}[t]{\textwidth}
            \centering
            \includegraphics[width=0.7\textwidth]{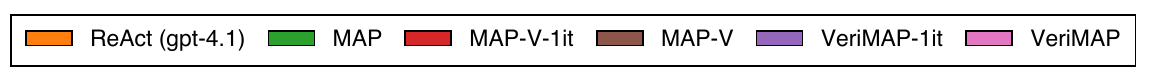}
        \end{subfigure}\\\vspace{-1mm}
    \begin{subfigure}[t]{0.19\textwidth}
        \includegraphics[width=\textwidth]{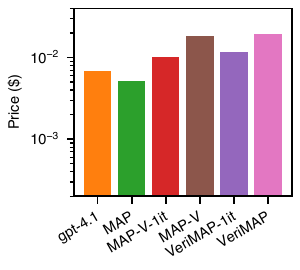}
        \scriptsize{\caption{MultiHopRAG}}
        \label{fig:multihoprag-cost}
    \end{subfigure}
    \hfill
    \begin{subfigure}[t]{0.19\textwidth}
        \includegraphics[width=\textwidth]{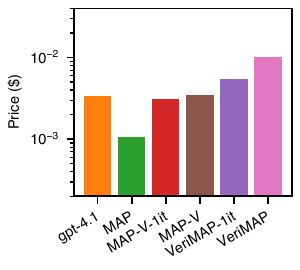}
        \scriptsize{\caption{HumanEval}}
        \label{fig:humaneval-cost}
    \end{subfigure}
    \hfill
    \begin{subfigure}[t]{0.19\textwidth}
        \includegraphics[width=\textwidth]{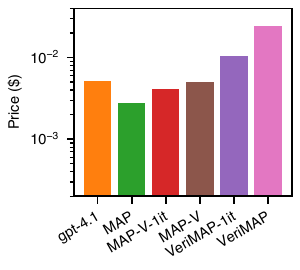}
        \scriptsize{\caption{BigCodeBench-Hard}}
        \label{fig:bigcodebench-cost} 
    \end{subfigure}
    \hfill
    \begin{subfigure}[t]{0.19\textwidth}
        \includegraphics[width=\textwidth]{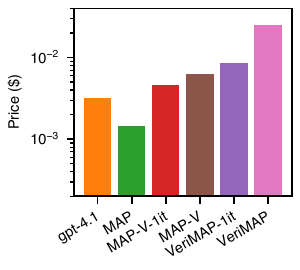}
        \scriptsize{\caption{GSM8K}}
        \label{fig:gsm8k-cost}
    \end{subfigure}
        \hfill
    \begin{subfigure}[t]{0.19\textwidth}
        \includegraphics[width=\textwidth]{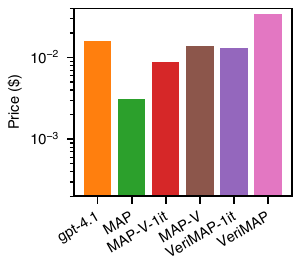}
        \scriptsize{\caption{Olympiads}}
        \label{fig:olympiads-cost}
    \end{subfigure}

    \caption{Total API cost (planner, executor, verifier) across five benchmarks. The y-axis indicates average per-task cost in USD, using \texttt{gpt-4.1} as the planner and \texttt{gpt-4o-mini} as the executor and verifier (when applicable).}
    \vspace{-1em}
    \label{fig:cost-analysis}

\end{figure*}
Figure~\ref{fig:cost-analysis} reports the average LLM API cost per task across benchmarks and methods, using \texttt{gpt-4.1} as the planner (Full calculation details in \Cref{app:cost-analysi}). Compared to the single-agent (ReAct) baseline, \sys generally incurs higher costs due to the added verification and retry mechanisms. However, since most of the reasoning is handled by the planner, \sys can delegate execution to less costly models, partially offsetting the overhead. Despite a modest increase in cost (on the order of \$0.001), these additional computations directly enable the performance gains reported in Table~\ref{tab:main_results}.

We further observe that the cost gap between \sys and the next-best performing method, \generic, depends on task complexity. For simpler tasks such as GSM8K, which likely overlap with model pre-training data or require limited reasoning, models can solve them efficiently, resulting in a larger cost gap where \sys’s additional verification offers limited utility. Conversely, for harder benchmarks like Olympiads, where models cannot easily derive efficient solutions, the cost gap narrows as \sys’s verification and replanning become essential for achieving correct results. This balance between cost and utility highlights the importance of selecting systems according to task difficulty and performance needs.
Further analysis about cost breakdown is in \Cref{app:cost-breakdown}.

\subsection{Verification Function Analysis}
\paragraph{VF statistics}
\begin{table}[t]
\centering
\footnotesize
\setlength{\tabcolsep}{5.5pt}
\resizebox{\columnwidth}{!}{
\begin{tabular}{l rr rr}
\toprule
\textbf{Dataset} &
\multicolumn{2}{c}{\textbf{Average Number}} &
\multicolumn{2}{c}{\textbf{Average Length}} \\
\cmidrule(lr){2-3}\cmidrule(lr){4-5}
 & \textbf{Python VF} & \textbf{NL VF}
 & \textbf{Python VF} & \textbf{NL VF} \\
\midrule
MultiHopRAG         & 0.25  & 7.36 & 159.7 & 190.0 \\
HumanEval           & 8.71  & 1.75 & 283.8 & 161.3 \\
BigCodeBench-Hard   & 14.15 & 2.41 & 381.6 & 174.3 \\
GSM8K               & 6.39  & 1.46 & 139.8 & 156.5 \\
Olympiads           & 10.96 & 5.07 & 243.8 & 165.0 \\
\bottomrule
\end{tabular}
}
\caption{Verification function profiling across datasets. Average Number refers to the average number of VF calls per task. Length is in characters.}
\label{tab:verification_profiling}
\vspace{-2.1em}
\end{table}
Table~\ref{tab:verification_profiling} presents the distribution and characteristics of planner-generated VFs from \sys. The planner adapts the VF type to the task domain: for programming benchmarks such as BigCodeBench-Hard, it predominantly produces Python VFs, averaging 14.15 executable checks per task, whereas for text-based tasks like MultiHopRAG, it primarily relies on LLM-based VFs (7.36 per task) to assess semantic correctness. The planner also varies the complexity of VFs with task difficulty, as reflected in the average VF length, 139.8 for GSM8K compared to 381.6 for BigCodeBench-Hard. These patterns suggest that the planner adjusts both the form and complexity of verification according to the nature of the task.

\paragraph{Error Analysis of Verification Functions}

\begin{table}[t]
\centering
\small
\resizebox{\columnwidth}{!}{
\begin{tabular}{@{}lcccc@{}}
\toprule
\textbf{Method}   & \multicolumn{2}{c}{\sys} & \multicolumn{2}{c}{\generic} \\
\cmidrule(l){2-3} \cmidrule(l){4-5}
                  & \multicolumn{1}{c}{FP} & \multicolumn{1}{c}{FN} & \multicolumn{1}{c}{FP} & \multicolumn{1}{c}{FN} \\
\midrule
MultiHopRAG       & 21.24\% & 4.81\%  & 18.80\% & 14.20\% \\
HumanEval         & 2.14\%  & 18.57\% & 13.51\% & 0.68\%  \\
BigCodeBench-Hard & 22.97\% & 21.62\% & 71.62\% & 0.00\%  \\
GSM8K             & 6.40\%  & 0.40\%  & 12.80\% & 1.80\%  \\
Olympiads         & 55.40\% & 1.00\%  & 65.40\% & 1.40\%  \\
\bottomrule
\end{tabular}
}
\caption{Error analysis of \sys and \generic on five benchmarks. FP stands for false positive, and FN stands for false negative.}\label{tab:fp-fn}
\vspace{-5mm}
\end{table}

Ideally, verification functions should neither incorrectly mark failed executions as passing (false positives) nor incorrectly flag correct executions as failing (false negatives). Table~\ref{tab:fp-fn} reports the frequency of false positives and false negatives for generic verifications from \generic and planner-generated verifications from \sys. \generic verifications tend to have low false negative rates due to their flexibility, but this same flexibility often leads to high false positive rates, indicating that the functions can be “too loose.” In contrast, \sys generally exhibits lower false positive rates\footnote{Except for MultiHopRAG, where most VFs are NL-based}, reflecting a more structured and comprehensive verification strategy. However, this occasionally comes at the cost of higher false negatives, particularly in HumanEval and BigCodeBench-Hard, where the approach can be “overly conservative.” Overall, even without external signals, \sys is able to generate VFs that more effectively monitor the execution of multi-agent systems.

\vspace{-2mm}
\subsection{Case study (Example from Olympiads)} 

\begin{tcolorbox}[title=Example Task, myprompt, breakable]
    Given that $(1+a)x^4+x^3-(3a+2)x^2-4a=0$, does there exist a $x_0\in \mathbb{R}$ that there is never a solution for any $a\in \mathbb{R}$?
\end{tcolorbox}
\vspace{3mm}
\begin{figure*}[ht]
    \centering
    \includegraphics[width=\textwidth]{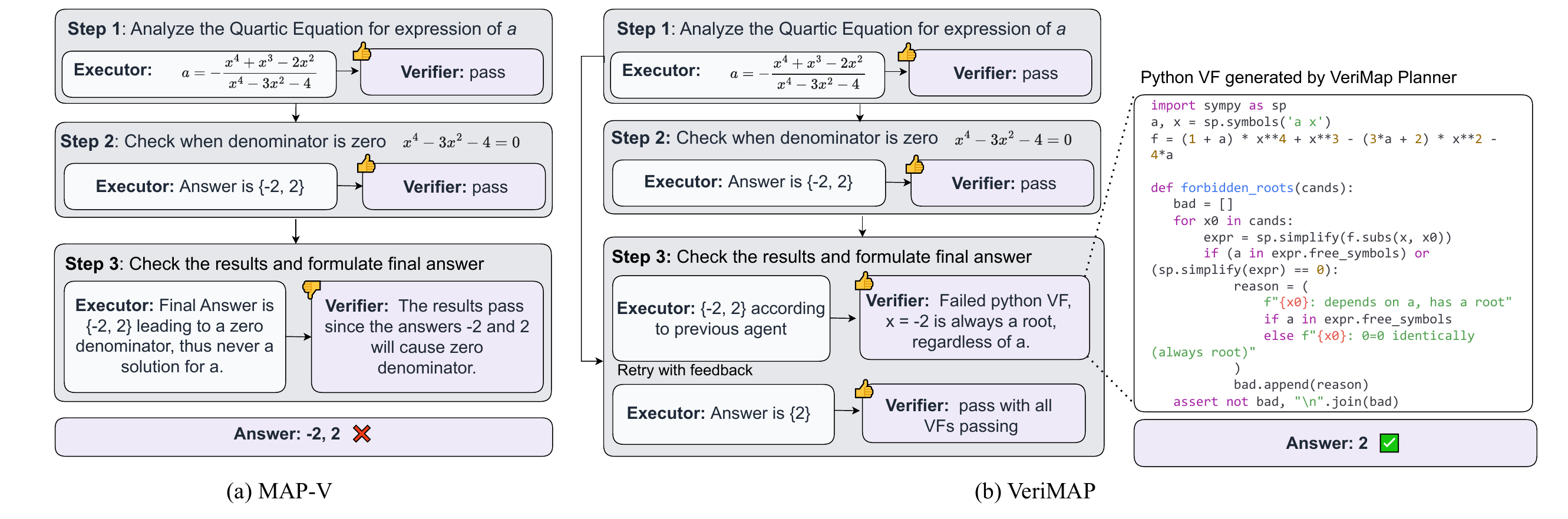}
    \caption{Comparison of planning and execution–verification traces between \generic and \sys. At step 3, the \generic verifier failed to detect the incorrect answer $-2$ (marked with a thumbs-down), whereas \sys, guided by the planner-generated VF, successfully identified the error and used the feedback to guide the executor to the correct answer in the next attempt.}
    \label{fig:case-study}
 
\end{figure*}
\vspace{-5mm}
Comparison between \generic and \sys solutions are shown in \Cref{fig:case-study}. Both reached a reasonable plan and arrived at a seemingly correct execution at step 2, identifying that $a=-\frac{x^4+x^3-2x^2}{x^4-3x^2-4}$, and that the denominator must be zero for $a$ to have no solution, yielding $x=\pm 2$. However, $(x+2)$ is a common factor of both the numerator and denominator. When $x=-2$, the equation always holds regardless of the value of $a$. Thus, $-2$ should not be a valid answer. 

With the generic natural language–based verification using only the context from step 2, \generic failed to verify the answer against either the numerator and the full expression, leading to a false positive verification result. In contrast, in \sys, even though the executor made the same initial mistake, Step 3 included (i) necessary context from Step 1 (see the connecting edge) and (ii) a structured Python-based verification that checks for \texttt{free\_symbols}. This allowed \sys to correctly capture the error, provide feedback to the executor, and ultimately converge on the correct solution.

This comparison highlights the effectiveness of planner-generated verifications in capturing nuanced and complex errors. Moreover, the subtle but necessary structural differences in the plan demonstrate how enforcing verification generation, in turn, improves overall plan quality.

\vspace{-1mm}
\section{Related Work}
\vspace{-2mm}

\paragraph{Planning with Language Models}
Planning has become a central focus in enabling LLMs to perform complex reasoning tasks, usually by decomposing them into intermediate and actionable steps. \citet{yao2023react} introduced ReAct, which interleaves reasoning and action to improve transparency and adaptability, though it remains limited on more complex tasks. ADaPT \cite{prasad-etal-2024-adapt} extends this idea by adaptively deciding the level of decomposition, enabling more flexible planning across different task types. Recent work emphasizes more structured frameworks, such as Plan-and-Act \cite{erdogan2025planandactimprovingplanningagents} which proposes a two-stage system with fine-tuned planners and executors, while \citet{kim-etal-2024-rada} use retrieval-augmented decomposition and trajectory exemplars. ExploraCoder~\cite{wang-etal-2025-exploracoder} distinguishes reactive, preactive, and exploratory strategies, showing exploratory hybrids to be more robust with unseen APIs in program synthesis. These systems highlight the promise of single-agent planning but also expose their susceptibility to compounding reasoning errors.

\paragraph{Multi-Agent Systems}
Unlike single-agent planning, multi-agent systems (MAS) distribute responsibilities across collaborating or competing agents, enabling specialization but introducing new challenges in coordination. Benchmarks such as MultiAgentBench \cite{zhu-etal-2025-multiagentbench} provide systematic evaluations of communication topologies and coordination protocols, showing how explicit planning can improve collective outcomes. Architectures such as MALMM \cite{singh2024malmmmultiagentlargelanguage} introduce hierarchical roles with supervisory agents to balance efficiency and adaptability.
RopMura~\citep{wu2025talkrightspecialistsrouting} employs a router to dynamically select agents and planners to solve a query, enabling accurate and efficient question answering.
Despite these advances, failures remain common. \citet{cemri2025multiagentllmsystemsfail} analyze why multi-agent LLM systems often underperform, pointing to unstable communication, misaligned objectives, and cascading coordination errors. These findings highlight that while MAS can mitigate single-agent limitations through distributed reasoning, they also introduce new challenges that demand stronger mechanisms for coordination, alignment, and reliability.

\paragraph{Verification}
Verification has emerged as a complementary component to planning. LLM-based evaluators include AutoCalibrate \cite{liu-etal-2024-calibrating}, which aligns evaluators to human labels, and EvoPrompt \cite{guo2024connecting}, which evolves prompts to improve evaluator reliability. \citet{stechly2024selfverificationlimitationslargelanguage} further show that LLMs struggle with reliable self-verification in reasoning and planning tasks, underscoring broader concerns about the stability of LLM evaluators \cite{10.1145/3654777.3676450}. In collaborative settings, planners may generate verifier agents \cite{suzgun2024metaprompting}, or frameworks may define verification steps automatically \cite{hao-etal-2025-large, grigorev2024verifyllm}. Formal approaches such as AdaPlanner~\cite{sun2023adaplanner} introduce symbolic verification, offering guarantees at the cost of flexibility. Recent work such as VeriLA \cite{sung2025verilahumancenteredevaluationframework} emphasizes human-centered evaluation for interpretable verification of agent failures, highlighting how interpretability and reliability remain open challenges. Together, these studies suggest verification is essential but often treated as an external module, raising questions about tighter integration with planning processes. An exhaustive comparison of existing systems with \sys{} is shown in~\Cref{tab:mas_verification_comparison}.
\vspace{-2.5mm}
\section{Conclusion}
\vspace{-2mm}
We introduced \sys{}, a framework for multi-agent collaboration with verification-aware planning. Planner-generated verification functions monitor and guide agent execution, enabling reliable coordination and self-refinement. Experiments show that \sys{} improves robustness, interpretability, and performance across diverse complex tasks.

\section*{Limitations}

While \sys demonstrates consistent improvements across diverse benchmarks, it is not without limitations. 

\paragraph{Replanning}
In the current \sys, when a node fails verification after all retries, the coordinator collects all execution traces and augments the original planning prompt with this history to generate a revised plan. The iterative refinement is essential to \sys as shown in its comparison with \sysOneR. While effective, this simple mechanism calls for further improvements. A key question is what signals to collect from failed executions and how to incorporate them effectively. Future work could move beyond appending the full history and instead analyze specific failure signals, such as error types or patterns across retries,to enable more targeted plan corrections, potentially improving both efficiency and robustness.

\paragraph{Centralized Planning}
\sys relies on a centralized planner that dictates task decomposition and context sharing. This makes the system heavily reliant on the capability of the planner model: weaker planners may generate suboptimal task decompositions or verification functions, leading to degraded performance. Although explicit and planner-driven coordination is generally more effective for tasks with well-specified requirements, there exists a more decentralized, “group-chat” style of planning where agents operate with greater autonomy under only lightweight guidance. This may be better suited for more creative or open-ended tasks. We leave the exploration of such decentralized approaches to future work.

\paragraph{Resource Constraints}
\sys{} currently assumes reliable access to multiple large models and external tools, which may limit its applicability in resource-constrained settings. Future work could focus on reducing cost through adaptive verification strategies, improving planner robustness, and adapting \sys to resource-constrained settings.

\section*{Ethical Considerations}
Our work does not involve sensitive personal data or human subjects. However, as with any system that relies on large language models, there are potential ethical risks. First, reliance on powerful LLMs may perpetuate biases embedded in training data, and our framework does not explicitly mitigate such biases. Second, the verification process, while designed to improve robustness, could inadvertently reinforce flawed reasoning if verification functions are poorly specified. Third, multi-agent coordination introduces additional complexity for transparency: while agents can correct each other, the planning and verification chain may reduce interpretability for end users. Finally, increased computational demand raises sustainability concerns due to energy consumption. We encourage future work to investigate bias-aware verification functions, develop more interpretable coordination mechanisms, and design energy-efficient verification strategies.
\paragraph{Disclosures on AI Usage}
The authors confirm that the majority of the text in this paper was written by the authors themselves. We used GPT-5 to assist with brainstorming ideas, editing for clarity, and improving language fluency, and Copilot to assist with coding. Any content generated by AI tools has been reviewed and edited by the authors, and we take full responsibility for the final content of the paper.

\bibliography{custom}

\appendix

\section{Prompts}

\subsection{\sys Planner Prompts}\label{sec:planner_prompts}
Figure~\ref{fig:planner-prompt} provides the detailed prompt used for the \sys~planner, where \{available\_tools\} is a placeholder for the list of tools the executors and verifiers can use, \{task\_instruction\} is the specific user task to be solved, and \{demo\_example\} is an optional few-shot task-specific demonstration of a \sys-generated plan.

Figure~\ref{fig:planner-replan-prompt} provides the replanning prompt inserted into the planning prompt when the initial plan fails, where \{previous\_dag\_str\} is the JSON representation of the previous failed DAG-based plan, and \{failed\_context\} is the traceback of the input/output of the failed node and two generations of ancestors of the failed node in the failed execution.

\begin{figure*}[!htb]
\begin{tcolorbox}[myprompt, breakable, title=Prompt for \sys~planner, width=\textwidth, fontupper=\tiny]
You are a planner responsible for generating high-level action plans to solve arbitrary tasks.

\textbf{Objective} \\
Decompose the input task into smaller, manageable \textbf{subtasks} that can be executed by \textbf{agents capable of calling external tools}.  
Tools available are \{available\_tools\}

Each subtask should:
\begin{itemize}[leftmargin=1.5em]
  \item Be represented as a \textbf{node} in a dependency graph.
  \item Contain a \textbf{high-level name} describing the subtask.
  \item Include a \textbf{detailed, executable instruction} for the agent, specifying how to perform the subtask and how to \textbf{utilize context from its dependencies}. You \textbf{MUST} specifically mention how to use the given contexts!
  \item Instructions for agents should focus on the goal of the subtask, the required inputs, and the expected output format. Avoid being overly prescriptive about the internal execution steps unless a specific algorithm is essential. Trust the agent's capabilities.
  \item Instructions should briefly explain the reasoning for the subtask. Why is this step necessary? What does it achieve for the overall plan? This provides crucial context.
  \item Each subtask instruction should be self-contained and self-explanatory: given the appropriate context or the output of previous nodes, another agent should be able to complete the subtask without needing to understand the overall task.
  \item Choose the granularity of subtasks carefully, ensuring that each subtask is solvable by an agent equipped with tools. If the user task is very simple, it is acceptable to generate a plan with only a single subtask.
\end{itemize}

\textbf{Verification Instruction} \\
Each subtask must include a set of verification functions to evaluate whether the agent has successfully completed the assigned task. When designing subtasks and instructions, ensure they are verifiable based on the provided context or input. Verification is supported in two main forms:
\begin{itemize}[leftmargin=1.5em]
  \item Python unit tests – Executed using a Python interpreter to perform structural and functional checks.
  \item LLM-based criteria – Judged by a tool-enabled LLM with access to the same tools as the agent.
\end{itemize}
Design subtasks with verifiability in mind to enable clear, objective evaluation of agent performance.

\textbf{Structure} \\
Represent the plan as a \textbf{Directed Acyclic Graph (DAG)} where:
\begin{itemize}[leftmargin=1.5em]
  \item Each \textbf{node} includes:
  \begin{itemize}
    \item \texttt{id}: a unique identifier
    \item \texttt{name}: a concise, high-level description of the subtask
    \item \texttt{instruction}: a detailed instruction tailored to the agent, including how to incorporate inputs from incoming nodes. \\[0.5em]

    \texttt{input}: list of required input variables, in the format of \texttt{<node.var>} the variable name \texttt{var} should match the output of its depending nodes. Empty list if it does not depend on the output of any previous node. Use \texttt{USER\_TASK} to refer to the raw user-provided input of the entire task. \\[0.5em]

    \texttt{output}: expected output variables \\[0.5em]

    \texttt{verification}: a JSON array of test specifications that the agent will run or evaluate to confirm correctness. Each element must include:
    \begin{itemize}
      \item \texttt{name} (string): a unique identifier for the test, e.g. \texttt{"test\_<subtask>\_<condition>"}.
      \item \texttt{type} (string): either \texttt{"python"} for executable assertions or \texttt{"llm"} for human-readable judge criteria.
      \item If \texttt{type == "python"}:
      \begin{itemize}
        \item \texttt{code} (string): one or more lines of self-contained Python, using the \texttt{inputs} and \texttt{outputs} dictionaries to assert correctness.
        \item Must cover at least:
        \begin{enumerate}
          \item Structural validation (e.g. output exists, correct type/format).
          \item Functional validation (e.g. correct mathematical result or string pattern).
          \item Edge cases (e.g. negative numbers, empty lists).
        \end{enumerate}
      \end{itemize}
      \item If \texttt{type == "llm"}:
      \begin{itemize}
        \item \texttt{content} (string): a clear instruction for an LLM ``judge'' to verify a particular requirement in the human-readable output (e.g. ``Ensure the summary mentions the terms `charges' or `indictment'").
      \end{itemize}
    \end{itemize}
    Tests should be fine-grained (one assertion per \texttt{code} block) and collectively ensure the subtask’s output both syntactically and semantically meets the instruction.
  \end{itemize}
  \item Each \textbf{edge} represents a \textbf{dependency or collaboration requirement} between nodes (e.g., information exchange, sequencing, coordination between agents).
\end{itemize}

\textbf{Formatting Instructions} \\
\noindent Output format (JSON):
\begin{verbatim}
{
  "nodes": [
    {
      "id": "node_id",
      "name": "subtask name",
      "instruction": "detailed and context-aware instruction for the agent",
      ...
    },
    ...
  ],
  "edges": [
    ["from_node_id", "to_node_id"],
    ...
  ]
}
\end{verbatim}
\begin{itemize}[leftmargin=1.5em]
  \item Do \textbf{not} include any additional text or explanations.
  \item Do \textbf{not} wrap the JSON output in code blocks or markdown formatting.
\end{itemize}
\textbf{Task Instructions} \\
\{task\_instruction\} \{demo\_example\}
\end{tcolorbox}
\vspace{-3mm}
\caption{Prompt for \sys Planner}\label{fig:planner-prompt}
\end{figure*}

\begin{figure*}[!htb]
\begin{tcolorbox}[myprompt, breakable, title=Replanning Prompt for \sys Planner, width=\textwidth, fontupper=\scriptsize]
The previous attempt to solve this task failed. Please generate a new plan based on the following guidance and context:

Previous DAG that failed:
\{previous\_dag\_str\}

Context of previous failure: 
\{failed\_context\}

Generate a revised DAG plan. Ensure the new plan addresses the previous issues.
\end{tcolorbox}
\caption{Replanning Prompt for \sys Planner}\label{fig:planner-replan-prompt}
\end{figure*}

\subsection{\sys Executor Prompts}\label{sec:executor_prompts}
Figure~\ref{fig:executor-sys-prompt} and Figure~\ref{fig:executor-task-prompt} provide the prompts used for the \sys~executor. It includes two parts: the system prompt that sets up the agent's behavior (Figure~\ref{fig:executor-sys-prompt}) and the task prompt (Figure~\ref{fig:executor-task-prompt}) that provides specific instructions for each subtask. In the system prompt, \{tool\_desc\} is a placeholder for the list of tools the agent can use.

\begin{figure*}[!htb]
\begin{tcolorbox}[myprompt, breakable, title=System Prompt for \sys~Executor, width=\textwidth, fontupper=\scriptsize]
You are designed to help with a variety of tasks, from answering questions to providing summaries to other types of analyses—and to prepare your outputs so that any downstream “child” agents can pick up and continue work seamlessly.

\textbf{Tools}

You have access to a wide variety of tools. You are responsible for using the tools in any sequence you deem appropriate to complete the task at hand. This may require breaking the task into subtasks and using different tools to complete each subtask.

You have access to the following tools:
\{tool\_desc\}

\textbf{Coordination with Other Agents}

You need to anticipate downstream needs because you are a part of a multi-agent system. Structure your output so that each child agent can identify what to do next—e.g. by providing context summaries, data snippets, or explicit instructions.

You may also receive additional contexts from upstream agents. These contexts may include relevant information, redundant details, or noise. \textbf{Use only relevant information} from the provided context; do not copy or summarize irrelevant content.

\textbf{Do not repeat work} already handled by upstream agents.

\textbf{Output Format}

Please answer in the same language as the question and use the following format:

\begin{verbatim}
Thought: The current language of the user is: (user's language). I need to use a tool to help me answer the question.
Action: tool name (one of tool_names) if using a tool.
Action Input: the input to the tool, in a JSON format representing the kwargs (e.g. {"input": "hello world", "num_beams": 5})
\end{verbatim}

Please ALWAYS start with a Thought.

NEVER surround your response with markdown code markers. You may use code markers within your response if you need to.

Please use a valid JSON format for the Action Input. If you include the "Action:" line, then you MUST include the "Action Input:" line too, even if the tool does not need kwargs, in that case you MUST use "Action Input: \{\}".

If this format is used, the tool will respond in the following format:

\begin{verbatim}
Observation: tool response
\end{verbatim}

You should keep repeating the above 
format till you have enough 
information to answer the question 
without using any more tools. At 
that point, you MUST respond in one 
of the following two formats:

\begin{verbatim}
Thought: I can answer without using any more tools. I'll use the user's language to answer
Answer: [your answer here (In the same language as the user's question)]
\end{verbatim}

\begin{verbatim}
Thought: I cannot answer the question with the provided tools.
Answer: [your answer here (In the same language as the user's question)]
\end{verbatim}

MAKE SURE that you include EVERYTHING needed in the answer part, as the verifier and downstream agents won't be able to see your thinking process, even though that might make the answer longer, that is necessary.

\textbf{Current Conversation}

Below is the current conversation consisting of interleaving human and assistant messages.
\end{tcolorbox}
\caption{System Prompt for \sys Executor}\label{fig:executor-sys-prompt}
\end{figure*}

In the task prompt, \{subtask\} is the name of the subtask to be executed, \{instruction\} is the detailed instruction for the subtask provided by the planner, and \{contexts\} are the relevant contexts provided by the planner in the structured input format.

\begin{figure*}[!htb]
\begin{tcolorbox}[myprompt, breakable, title=Task Prompt for \sys~Executor, width=\textwidth, fontupper=\small]
You are working on a subtask, which is part of a complete task.

\textbf{Input}

\textbf{Your task:}
\{subtask\}

\textbf{Instructions for your subtask:}
\{instruction\}

\textbf{Contexts for your subtask:}
\{contexts\}

\textbf{Structure}

Please first go through the Thought-Action process.

\textbf{Instructions}

\begin{itemize}
\item Do \textbf{not} include any additional text or explanations.
\item Do \textbf{not} wrap the JSON output in code blocks or markdown formatting.
\item Do \textbf{not} output again the observations. They are provided to you, don't repeat them.
\item Do \textbf{not} put any symbols like "," "'" after your "Action Input:"'s json. The only thing after your "Action Input" should be a valid json.
\end{itemize}

\end{tcolorbox}
\caption{Task Prompt for \sys Executor}\label{fig:executor-task-prompt}
\end{figure*}

\subsection{\sys Verifier Prompts}\label{sec:verifier_prompts}
Figure~\ref{fig:verifier-prompt} provides the prompt used for the \sys~verifier for LLM-based verifications, where \{agent\_input\} is the original task input, \{verify\_prompt\} is the verification requirement provided by the planner, and \{agent\_output\} is the structured output produced by the executor.

\begin{figure*}[!htb]
\begin{tcolorbox}[myprompt, breakable, title=Prompt for \sys~Verifier, width=\textwidth, fontupper=\small]
You are an expert evaluator charged with judging whether an agent’s response meets a given verification requirement, based on three pieces of information: the original task input, the verification requirement, and the agent’s output.

\textbf{Objective:}
Assess the correctness of the agent’s output against the verification requirement. Always leverage your tools to perform independent verification of any facts, data, or claims.

\textbf{Scoring Scale:}
\begin{description}
\item[\textbf{0 (Failure):}] Output misses key requirements or contains serious errors, or output is wrong in format.
\item[\textbf{1 (Success):}] Output adequately satisfies the requirement.
\end{description}

\textbf{Response Format (JSON only):}
\begin{verbatim}
{
"success_score": <0 or 1>,
"reasoning": "<1–2 sentence justification, explain IN DETAIL, SPECIFICALLY about how you will amend 
              the answer>"
}
\end{verbatim}

\textbf{Context Variables:}
\begin{itemize}
\item \textbf{Agent Input:} \{agent\_input\}

\item \textbf{Verification Requirement:} \{verify\_prompt\}

\item \textbf{Agent Output:} \{agent\_output\}
\end{itemize}

\textbf{Guidelines:}
\begin{enumerate}
\item Return \textit{only} the JSON object—no extra text, markdown, or code blocks.
\item \texttt{success\_score} must be exactly 0 or 1.
\item \textbf{MANDATORY:} Independently verify all claims using your available tools (e.g., fact‐checks, data lookups, calculation scripts). Never assume the agent’s explanation is complete or correct without tool-based confirmation.
\item Focus strictly on correctness relative to the requirement; level of explanatory detail in the agent’s output beyond correctness is \textit{not} required.
\end{enumerate}
\end{tcolorbox}
\caption{Prompt for \sys Verifier}\label{fig:verifier-prompt}
\end{figure*}
\section{Implementation Details}\label{sec:implementation}

\subsection{Dataset Details}\label{sec:dataset_details}
\begin{table*}[t]
\centering
\small
\begin{tabular}{l m{9.5cm} l l}
\hline
\textbf{Dataset} & \textbf{Description} & \textbf{\# Training} & \textbf{\# Test} \\
\hline
MultiHopRAG* & Multi-hop QA over a news knowledge base~\citep{tang2024multihoprag} & 500 & 500 \\
\hline
HumanEval & Python function synthesis from docstrings; evaluated by unit tests~\citep{chen2021evaluating} & 164 & 164 \\
\cline{1-4}
BigCodeBench-Hard & Execution-based code generation with diverse library calls and complex instructions; evaluated by unit tests~\citep{zhuo2024bigcodebench} & 74 & 74 \\
\hline
GSM8K* & Grade-school math word problems requiring multi-step reasoning~\citep{cobbe2021training} & 500 & 500 \\
\cline{1-4}
Olympiads* & Competition math problems with standardized chain-of-thought solutions~\citep{li2024numinamath} & 500 & 500 \\
\hline
\end{tabular}
\caption{Dataset statistics. * represents samples of the original datasets.}\label{tab:datasets}
\end{table*}

A detailed description of the datasets used in our experiments is provided in Table~\ref{tab:datasets}. For all datasets, we choose at most 500 random samples as the training set and at most 500 random samples as the test set. For datasets with fewer than 500 samples, we use half of the samples as the training set and the other half as the test set. The training set is only used for few-shot learning and Training-based \sys, while the test set is used for evaluation.

Across the five datasets, MultiHopRAG, HumanEval, BigCodeBench-Hard, and GSM8K are used as-is. The Olympiads dataset is originally from the NuminaMath-CoT benchmark~\citep{li2024numinamath}, which contains 32,926 problems. We filter out competition math problems with standardized chain-of-thought solutions, resulting in 1,337 problems. We then randomly select 500 problems as the training set and the other 500 problems as the test set. The full dataset is available at \url{https://huggingface.co/datasets/Metaskepsis/Olympiads}.

\subsection{Executor Implementation Details}\label{sec:executor_implementation}
In our implementation, the Executor is a \emph{ReAct}-style single agent. It is provided with a ReAct system prompt, the task name, detailed instructions, and relevant contexts from the Coordinator. The system prompt explicitly directs the agent to produce its final answer in a structured output format, and the task prompt provides a detailed description of the subtask along with a structured output guide. The structured output guide specifies the expected output variables, their types, and any constraints or formats that need to be adhered to. An example of the ReAct system prompt and task prompt is provided in Appendix~\ref{sec:executor_prompts}.

The agent is allowed a maximum of 20 rounds to complete a given subtask. If the agent fails to produce a final, structured output within these 20 rounds, the execution is considered a failure, and the process proceeds directly to the error handling stage.

\subsection{Verifier Implementation Details}\label{sec:verifier_implementation}
In our implementation, each executor node can be equipped with a suite of verification functions. A subtask is deemed successfully completed only when its output passes all associated verification functions. If any single verifier fails, the execution is considered a failure, and the process is escalated to the Coordinator's error handling stage.

The feedback mechanism is tailored to the type of verification being performed:

\begin{enumerate}
\item For \textbf{Python-based verification}, if a programmatic assertion fails, the system automatically captures and returns the complete traceback and the specific assertion message as feedback. This provides precise, actionable information for debugging and correction.

\item For \textbf{LLM-based verification}, the verifier LLM is set to be the same as the executor LLM. Similarly, it is powered by a ReAct-style agent and has access to the same set of tools as executors. It is prompted to produce a structured output containing a binary success/fail evaluation and a detailed textual feedback given the LLM-based verification prompt and the output from the executor. This feedback is designed to be constructive, explicitly identifying which parts of the original answer were incorrect and providing a comprehensive solution for how to amend it. Prompts used for LLM-based verification are provided in Appendix~\ref{sec:verifier_prompts}.
\end{enumerate}

\subsection{Coordinator Implementation Details}\label{sec:coordinator_implementation}

\begin{algorithm*}[t]
\caption{Task Coordinator for Multi-Agent Execution with Multiple Verifiers per Node}
\label{alg:task-coordinator-multi-vf}
\begin{algorithmic}[1]
\Require $P = (V, E)$ \Comment{Task plan as DAG with nodes $V$ and edges $E$}
\Require $\{VF_v \mid v \in V\}$, where $VF_v = \{VF_{v,1}, VF_{v,2}, \dots\}$ \Comment{Node-specific verification functions}
\Require $R_{\max}$ \Comment{Maximum retry limit per node}
\Require $I_{\max}$ \Comment{Maximum replanning iteration limit}
\Ensure Result $\in \{\text{exec\_out}, \text{failure}\}$ with execution traces

\State $order \gets \text{TOPOLOGICAL-SORT}(P)$
\State $iter \gets 0$ \Comment{Replanning iteration counter}

\While{$iter < I_{\max}$}
  \For{each node $v \in order$}
    \State $r \gets 0$ \Comment{Retry counter}
    \Repeat
      \If{$r = 0$}
        \State $ctx \gets \text{COMPILE-CONTEXT}(v)$
      \Else
        \State $ctx \gets \text{COMPILE-CONTEXT}(v, exec\_out)$ \Comment{Use previous output}
      \EndIf
      \State $exec\_out \gets \text{EXECUTE}(v, ctx)$ \Comment{Invoke executor}
      
      \State $ver\_res \gets \text{VERIFY}(exec\_out, V    )$
\Comment{Invoke verifier}
      \If{$ver\_res = \text{pass}$}
        \State \textbf{break} \Comment{Proceed to next node}
      \Else
        \State $r \gets r + 1$
      \EndIf
    \Until{$ver\_res = \text{pass}$ \textbf{ or } $r \geq R_{\max}$}

    \If{$ver\_res \neq \text{pass}$}
      \State $traces \gets \text{COLLECT-TRACES}(v, exec\_out, ver\_res)$
      \State $P \gets \text{REPLAN}(P, traces)$ \Comment{Generate new plan}
      \State $iter \gets iter + 1$
      \State \textbf{goto line 1} \Comment{Start next iteration from new plan}
    \EndIf
  \EndFor
  \State \Return $exec\_out$ \Comment{Return latest execution result of last node}
\EndWhile

\State \Return failure

\end{algorithmic}
\end{algorithm*}

The Coordinator's orchestration logic is governed by specific parameters and data handling rules to ensure deterministic behavior.

\begin{itemize}[leftmargin=1.5em]
\item \textbf{Execution Model:} To simplify implementation, the Coordinator executes the nodes in the DAG sequentially, following a standard topological sort. Parallel execution is left for future work.
\item \textbf{Inter-node Data Flow:} When a node has multiple parent nodes, the Coordinator merges their structured outputs into a single dictionary. It then provides this dictionary as context to the child node's Executor, which can access the required inputs by their corresponding keys.
\item \textbf{Final Output:} The DAG contains one node explicitly marked as the final output node. The Coordinator returns only the structured output from this specific node upon successful completion of the entire graph.
\end{itemize}

\subsection{Baseline Implementation Details}
\label{sec:appendix_baselines}

\begin{table*}[!htb]
\centering
\small
\renewcommand{\arraystretch}{2.5}
\setlength{\tabcolsep}{3pt}
\begin{tabular}{
p{3cm} 
>{\centering\arraybackslash}p{2cm} 
>{\centering\arraybackslash}p{2cm} 
>{\centering\arraybackslash}p{2cm} 
>{\centering\arraybackslash}p{2cm} 
>{\centering\arraybackslash}p{2cm} 
>{\centering\arraybackslash}p{2cm} }
\toprule
\textbf{System} &
\textbf{MAS?} &
\textbf{Verification Entity} &
\textbf{Planner Generated Verification Module?} &
\textbf{Python Verification?} &
\textbf{Structured I/O?} &
\textbf{Retry?} \\
\midrule
\textbf{\sys{}} (ours) & \cmark & Subtask-level & \cmark & \cmark & \cmark & \cmark \\
\textbf{AdaPlanner}~\citep{sun2023adaplanner} & \xmark & Plan-level & \xmark & \makecell{\xmark \\\tiny prompt-based check} & \xmark & \cmark \\
\textbf{MetaPrompt}~\citep{suzgun2024metaprompting} & \makecell{\xmark \\\tiny single LM Meta Model} & Plan \& Execution level & \xmark & \cmark & \xmark & \xmark \\
\textbf{VerifyLLM}~\citep{grigorev2024verifyllm} & \xmark & Plan-level & \xmark & \xmark & \xmark & \xmark \\
\textbf{Self-Verification} \citep{stechly2024selfverificationlimitationslargelanguage} & \xmark & Plan-level & \xmark & \cmark & \xmark & \cmark \\
\textbf{Formal Verification Tools} \citep{hao-etal-2025-large} & \xmark & Plan \& Execution-level & \xmark & \makecell{\qmark \\\tiny SMT solver} & \makecell{\qmark \\ \tiny SMT format} & \cmark \\
\bottomrule
\end{tabular}
\caption{Comparison of planning and verification systems. 
\small
\textbf{MAS?}: Multi-Agent System; 
\textbf{Verification Entity}: Granularity of verification process; 
\textbf{Planner Generated Verification Module?}: Verification module generated by the planner; 
\textbf{Python Verification?}: Verification implemented in Python; 
\textbf{Structured I/O?}: System uses structured input/output; 
\textbf{Retry?}: Retry planning after verification or execution failure;  
\cmark: feature supported; 
\xmark: feature not supported; 
\qmark: partial/uncertain support.
 }
\label{tab:mas_verification_comparison}
\vspace{-3mm}
\end{table*}

\paragraph{\textbf{Single-Agent Baselines (ReAct)}} These agents utilize the ReAct framework and have access to the same set of tools as the \sys executor.
Their internal reasoning process is limited to a maximum of 20 steps per task.

\paragraph{\textbf{\noveri Data Handling}} In the \textbf{\noveri} ablation, where structured I/O is disabled, an executor node is provided with the complete prompt and raw text output from all of its parent nodes up to two generations back. This historical context helps compensate for the missing structured data. The prompts given to the executor are identical to those in the full \sys system, with the exception of clauses related to generating structured outputs and verification criteria.

\paragraph{\generic Verifier} The verifier in the \textbf{\generic} baseline is powered by the same LLM as the executor. Its implementation is consistent with a standard LLM-based verification, where its goal is to check for alignment between the generated output and the specific instructions the executor received for that subtask.

\paragraph{Single-Round Execution} In the \textbf{\genericOneR} and \textbf{\sysOneR} configurations, the system completes one full pass of the DAG. If a node's verification fails, its output (whether incorrect or empty) is still passed as context to any subsequent child nodes for the remainder of that single iteration. No re-execution or re-planning is triggered.

\subsection{Cost calculation}
\label{app:cost-analysi}
For \texttt{gpt-4o-mini} as the executor, we assume the cost per 1M tokens as \$0.15 for input, \$0.08 for cached input (inputs that have been seen before), and \$0.60 for output. For \texttt{gpt-4.1} as the planner, we assume the cost per 1M tokens as \$2.00 for input, \$0.50 for cached input, and \$8.00 for output.
\section{Training-based \sys}\label{sec:dpo}\label{sec:veriplan-dpo}

To further enhance the quality of plan generation, we fine-tuned the Verification-Aware Planner using Direct Preference Optimization (DPO)~\citep{rafailov2024directpreferenceoptimizationlanguage}. DPO is a methodology for aligning language models with human (or, in this case, machine-labeled) preferences. It operates on a dataset of triplets, each containing a prompt ($p$), a preferred response ($y_w$), and a less-preferred response ($y_l$). By training on these triplets, the model learns to increase the likelihood of generating responses similar to $y_w$ while decreasing the likelihood of those similar to $y_l$. Our process for creating this dataset and training the model is detailed below.

\subsection{Data Collection}
Our first step was to construct a dataset of preference pairs derived from the execution traces of \sys. We designated a portion of our datasets to serve as a training set for this purpose as in Appendix~\ref{sec:dataset_details}. On these tasks, we ran \sys using \texttt{gpt-4.1} as the Planner and \texttt{gpt-4o-mini} for both the Executor and Verifier roles. To capture a variety of outcomes, each task was executed three times.

From these runs, we collected all execution iterations. For a given task (the prompt, $p$), a plan that led to a successful, verified outcome—often after one or more error-correction cycles—was labeled as the \textit{preferred response} ($y_w$). Conversely, a plan generated within an iteration that ultimately failed verification and led to the re-planning stage was labeled as the \textit{less-preferred response} ($y_l$). This process yielded a raw dataset of $(p,y_w,y_l)$ triplets.

\subsection{Data Cleaning and Augmentation}
The raw dataset collected in the previous step had two primary limitations: 1) some failures may have been caused by random executor errors rather than genuinely poor plans, and 2) the diversity of planning errors was limited. We addressed these issues with a two-stage process.

First, to filter for meaningful preference pairs, each triplet was provided to \texttt{gpt-4.1} to judge whether the preferred and less-preferred plans exhibited a significant, structural difference in quality. Triplets where the plans were too similar, suggesting a non-deterministic execution failure, were removed from the dataset.

Second, to augment the dataset and improve its diversity, we first categorized common planning flaws observed in our experiments:
\begin{itemize}[leftmargin=1.5em]
    \item Boundary condition removed/changed (e.g., off-by-one, empty input)
    \item Incorrect parameter/default value (e.g., ddof, timeout)
    \item Missing or overly broad exception handling
    \item Data-type or casting mistake (e.g., int vs float)
    \item Variable naming shadow or mismatch
    \item Inefficient algorithm (e.g., $O(n^2)$ vs. a built-in function)
    \item Deleted necessary pre/post-processing (e.g., dropna(), .strip())
    \item Style/format only errors (e.g., extra header, comment)
\end{itemize}
We then used \texttt{gpt-4.1} to create new, synthetic training samples. For each high-quality triplet, we prompted the model to introduce one or more of the error types above into the plan to generate a new, even less-preferred version. This created additional, challenging negative examples for the DPO training.

\subsection{Hyperparameters and Training Details}
Our initial data collection process yielded 150 high-quality preference samples. After the cleaning and augmentation stages, the final dataset was expanded to 300 samples.

We used \texttt{gpt-4.1} as the base model for fine-tuning, which is the same model used for the Planner in our main experiments to ensure a fair comparison. The model was fine-tuned using the Direct Preference Optimization implementation available through OpenAI's fine-tuning API. The training was configured with the following hyperparameters: a \textbf{batch size} of 2, a single training \textbf{epoch}, a \textbf{learning rate multiplier} of 0.8, and a DPO \textbf{beta} ($\beta$) of 0.2.

\section{Additional Experiment Results}\label{sec:add_experiments}
\subsection{Iteration counts}
\label{app:iteration_counts}

\begin{table}[t]
\centering
\setlength{\tabcolsep}{4pt}
\begin{tabular}{lcc}
\toprule
 & \multicolumn{2}{c}{\textbf{Avg. Iter./Avg. Retr.}} \\
\cmidrule(lr){2-3}
 & \sys & \generic \\
\midrule
MultiHopRAG       & 1.48 / 1.57 & 1.60 / 1.78 \\
HumanEval         & 1.33 / 1.71 & 1.05 / 1.09 \\
BigCodeBench-Hard          & 1.85 / 2.28 & 1.07 / 1.11 \\
GSM8K             & 1.17 / 1.14 & 1.09 / 1.35 \\
Olympiads         & 1.84 / 2.29 & 1.46 / 1.54 \\
\bottomrule
\end{tabular}
\caption{Average iterations/retries of \sys vs.\ \generic across benchmarks.}
\label{tab:avg_iters}
\end{table}
\begin{figure*}[t]
    \centering
    \small
    
        \begin{subfigure}[t]{\textwidth}
            \centering
            \includegraphics[width=0.7\textwidth]{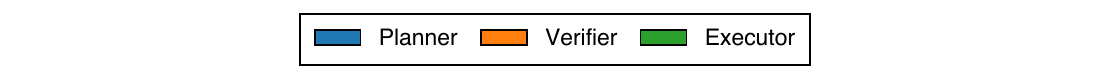}
        \end{subfigure}\\\vspace{-1mm}
    \begin{subfigure}[t]{0.19\textwidth}
        \includegraphics[width=\textwidth]{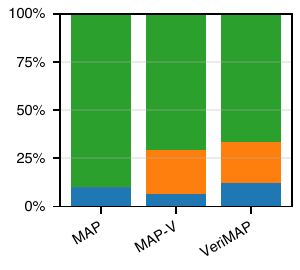}
        \scriptsize{\caption{MultiHopRAG}}
        \label{fig:multihoprag-cost-decomp}
    \end{subfigure}
    \hfill
    \begin{subfigure}[t]{0.19\textwidth}
        \includegraphics[width=\textwidth]{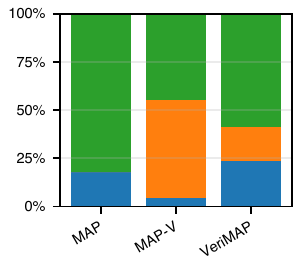}
        \scriptsize{\caption{HumanEval}}
        \label{fig:humaneval-cost-decomp}
    \end{subfigure}
    \hfill
    \begin{subfigure}[t]{0.19\textwidth}
        \includegraphics[width=\textwidth]{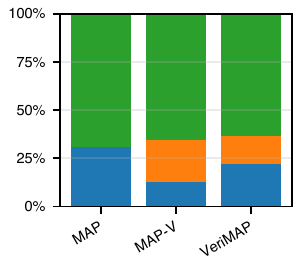}
        \scriptsize{\caption{BigCodeBench-Hard}}
        \label{fig:bigcodebench-cost-decomp} 
    \end{subfigure}
    \hfill
    \begin{subfigure}[t]{0.19\textwidth}
        \includegraphics[width=\textwidth]{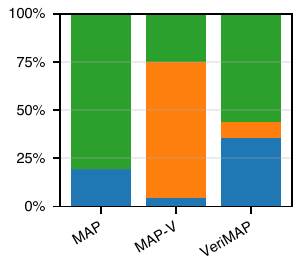}
        \scriptsize{\caption{GSM8K}}
        \label{fig:gsm8k-cost-decomp}
    \end{subfigure}
        \hfill
    \begin{subfigure}[t]{0.19\textwidth}
        \includegraphics[width=\textwidth]{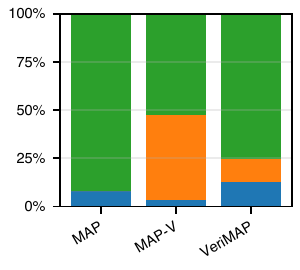}
        \scriptsize{\caption{Olympiads}}
        \label{fig:olympiads-cost-decomp}
    \end{subfigure}

    \caption{Cost decomposition of different components in \sys~and the \noveri and \generic baselines across all benchmarks. The cost is broken down into the planner cost, the executor cost, and the verifier cost.}
    \label{fig:cost-decomposition}

\end{figure*}

Table~\ref{tab:avg_iters} shows the average number of iterations and retries for \sys and the \generic baseline. Through more flexible and rigorous self-correction, \sys achieves superior accuracy. While on tasks with clearer success criteria like MultiHopRAG, \sys demonstrates greater efficiency with fewer iterations (1.48 vs. 1.60) and retries (1.57 vs. 1.78). On Programming and Olympiads tasks, \sys exhibits significantly higher average iterations and retries (e.g., 1.85 iterations and 2.28 retries on BigCodeBench-Hard, vs. 1.07 and 1.11 for the \generic baseline). We attribute the higher iteration counts to its effective verification mechanism. \sys's sophisticated verifiers successfully detect subtle flaws that the weaker \generic verifier misses. The \generic baseline often terminates prematurely after an undetected error, leading to lower iteration counts but also poor final accuracy. In contrast, \sys correctly identifies these failures and puts more effort in retries and re-planning to find a correct solution. Therefore, the higher iteration counts on difficult problems represent the necessary cost of achieving higher quality and robustness, directly explaining the performance gains reported in Table~\ref{tab:main_results}.

\subsection{Cost breakdown}
\label{app:cost-breakdown}
Figure~\ref{fig:cost-decomposition} shows the detailed cost breakdown of different components in \sys and the \noveri and \generic baselines across all benchmarks. A consistent cost distribution is evident across all methods, where the Executor is the most expensive component, followed by the Verifier, and finally the Planner. A core finding is that \sys's Planner cost (the blue bar) is consistently higher than that of the \generic baseline. This reflects the additional work required from our planner to generate not just an execution plan but also the sophisticated, structured verification logic. However, this upfront investment in planning leads to significant efficiency gains during the verification stage, particularly on programming tasks like HumanEval and BigCodeBench-Hard. On these benchmarks, \sys's Verifier cost (the orange bar) is substantially lower because it generates token-efficient, executable Python unit tests instead of relying on lengthy, prompt-based checks. This demonstrates that \sys's more intelligent planner directly enables a more effective and economical verification process.
\subsection{Training-based \sys}
We have experimented with the training based \sys-Planner as introduced in Section~\ref{sec:dpo}. It has achieved accuracies of 64.77\%, 93.06\%, 37.68\%, 93.00\% and 44.85\% on the MultiHopRAG, HumanEval, BigCodeBench-Hard, GSM8K, and Olympiads datasets, respectively. The training-based \sys-Planner achieves the best performance on the challenging Olympiads benchmark, with a significant +3.65\% improvement over the standard \sys. This indicates that fine-tuning the Planner to generate more robust and verifiable plans is particularly beneficial for extremely difficult tasks. However, on other benchmarks, \sys-DPO shows mixed results, sometimes underperforming compared to the original \sys, suggesting that while DPO training can enhance plan quality in certain contexts, it may also reduce generality in others. Further investigation is needed to fully understand the conditions under which DPO training provides consistent benefits.

\end{document}